\pdfoutput=1

\documentclass[11pt]{article}

\usepackage[]{acl}

\usepackage{times}
\usepackage{latexsym}
\usepackage{multirow}
\usepackage{helvet}  
\usepackage{courier}  
\usepackage{graphicx} 
\usepackage[T1]{fontenc}
\usepackage[utf8]{inputenc}
\usepackage{microtype}

\title{Aspect-specific Context Modeling for Aspect-based Sentiment Analysis}

\author{Fang Ma, Chen Zhang, Bo Zhang, Dawei Song\Thanks{ Dawei Song is the corresponding author.} \\
   School of Computer Science, Beijing Institute of Technology \\
   Beijing, China \\
   \texttt{\{mfang,czhang,bo.zhang,dwsong\}@bit.edu.cn}
}

\begin{document}
\maketitle
\begin{abstract}
Aspect-based sentiment analysis (ABSA) aims at predicting sentiment polarity (SC) or extracting opinion span (OE) expressed towards a given aspect. Previous work in ABSA mostly relies on rather complicated aspect-specific feature induction. Recently, pretrained language models (PLMs), e.g., BERT, have been used as context modeling layers to simplify the feature induction structures and achieve state-of-the-art performance. However, such PLM-based context modeling can be not that aspect-specific. Therefore, a key question is left under-explored: how the aspect-specific context can be better modeled through PLMs? To answer the question, we attempt to enhance aspect-specific context modeling with PLM in a non-intrusive manner. We propose three aspect-specific input transformations, namely aspect companion, aspect prompt, and aspect marker. Informed by these transformations, non-intrusive aspect-specific PLMs can be achieved to promote the PLM to pay more attention to the aspect-specific context in a sentence. Additionally, we craft an adversarial benchmark for ABSA (advABSA) to see how aspect-specific modeling can impact model robustness. Extensive experimental results on standard and adversarial benchmarks for SC and OE demonstrate the effectiveness and robustness of the proposed method, yielding new state-of-the-art performance on OE and competitive performance on SC.  \footnote{The code and proposed data are available at \url{https://github.com/BD-MF/ASCM4ABSA}.}
\end{abstract}

\section{Introduction}

\begin{figure}[t]
  \centering
  \includegraphics[width=\linewidth]{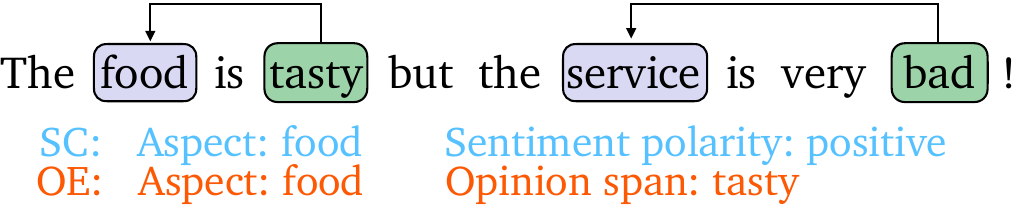}
  \caption{Example of the SC and OE. The words highlighted in purple represent the given aspects, whereas the words in green represent the corresponding opinion.}
  \label{absa_example}  
\end{figure}

Aspect-based sentiment analysis (ABSA) aims to infer multiple fine-grained sentiments  from the same content, with respect to multiple aspects. A fine-grained sentiment in ABSA can be categorized into two forms, i.e., sentiment and opinion. Accordingly, two sub-tasks of ABSA are aspect-based sentiment classification (SC for short) and aspect-based opinion extraction (OE for short). Given an aspect in a sentence, SC aims to predict its sentiment polarity, while OE aims to extract the corresponding opinion span expressed towards the given aspect. Figure~\ref{absa_example} shows an example of SC and OE. In the sentence ``\textit{The food is tasty but the service is very bad!}'', if \textit{food} is the given aspect, SC requires a model to give a \texttt{positive} sentiment on \textit{food} while OE requires a model to extract \textit{tasty} as the opinion span for the aspect \textit{food}.

An effective ABSA model typically would require either aspect-specific feature induction or context modeling. Prior work in ABSA largely relies on rather
complicated aspect-specific feature induction to achieve a good performance. Recently, pretrained language models (PLMs) have been shown to enhance the state-of-the-art ABSA models due to their extraordinary context modeling ability. However, currently the use of PLMs in these ABSA models is aspect-general, but overlooks two key questions: 1) whether the context modeling of a PLM can be aspect-specific; and 2) whether the aspect-specific context modeling within a PLM can further enhance ABSA.

To address the aforementioned key questions, in this paper, we propose to achieve \textit{aspect-specific context modeling} of PLMs with \textit{aspect-specific input transformations}. In addition to the commonly used aspect-specific input transformation that appends an aspect to a sentence, i.e., \textbf{aspect companion}, we  propose two more aspect-specific input transformations, namely \textbf{aspect prompt} and \textbf{aspect marker}, to explicitly mark a concerned aspect in a sentence. Aspect prompt shares a similar idea with aspect companion, except that it appends an aspect-oriented prompt instead of sole aspect description to the sentence. Aspect marker distinguishes itself from the above two by introducing two marker tokens, one before and the other after the aspect. As the proposed input transformations are intended to highlight a specific aspect, they in turn can be leveraged to promote the PLM to pay more attention to the context that is relevant to the aspect. Methodologically, this is achieved with a novel aspect-focused PLM fine-tuning model that is guided by the input transformations and essentially performs a joint context modeling and aspect-specific feature induction.

We conduct extensive experiments on both subtasks of ABSA, i.e., SC and OE, with various standard benchmarking datasets for effectiveness test, along with our crafted adversarial ones for robustness test. Since there are only datasets  for robustness tests in SC and is currently no dataset for robustness tests in OE, we propose an adversarial benchmark (advABSA) based on~\cite{xing2020tasty}'s datasets and methods. That is, the advABSA benchmark can be decomposed to two parts, where the first part is \textsc{ARTS-SC} for SC reused from~\cite{xing2020tasty} and the second part is \textsc{ARTS-OE} for OE crafted by us. The results show that models with aspect-specific context modeling achieve the state-of-the-art performance on OE and also outperform various strong SC baseline models without aspect-specific modeling. Overall, these results indicate that aspect-specific context modeling for PLMs can further enhance the performance of ABSA. 


To better understand the effectiveness of the three input transformations, we carry out a series of further analyses. After injecting aspect-specific input transformations into a sentence, we observe that the model attends to the correct opinion spans. Hence, we expect that a simple model with aspect-specific context modeling yet without needing complicated aspect-specific feature induction would serve as a sufficiently strong approach for ABSA.

\section{Related Work}

\subsection{Aspect-based Sentiment Classification SC}

ABSA falls in the broad scope of fine-grained opinion mining. As a sub-task of ABSA, SC determines the sentiment polarity of a given aspect in a sentence and has recently emerged as an active research area with lots of aspect-specific feature induction approaches. These approaches range from memory networks \citep{tang2016aspect, wang2018target}, convolutional networks \citep{li2018transformation, huang2018parameterized}, attentional networks \citep{wang2016attention, ma2017interactive}, to graph-based networks \citep{zhang2019aspect, zhang2019syntax, wang2020relational, tang2020dependency}. More recently, PLMs such as BERT~\citep{devlin2019bert} and RoBERTa~\citep{liu2019roberta}, have been applied to SC in a context-encoder scheme~\citep{yu2019adapting, li2019exploiting, xu2019bert,liang2019novel, song2020utilizing, yadav2021human} and achieved the state-of-the-art performance. However, PLMs in these models are aspect-general. We aim to achieve aspect-specific context modeling with PLMs so that these models can be further improved.

\subsection{Aspect-based Opinion Extraction OE}

OE is another sub-task of ABSA, first proposed by \citet{fan2019target}. It aims to extract from a sentence the corresponding opinion span describing an aspect. Most work in this area treats OE as a sequence tagging task, for which complex methods are developed to capture the interaction between the aspect and the context \citep{fan2019target, wu2020latent, feng2021target}. More recent models such as TSMSA-BERT~\citep{feng2021target} and ARGCN-BERT~\citep{jiang2021attention}, adopt PLMs. In TSMSA-BERT, the multi-head self-attention is utilized to enhance the BERT PLM. ARGCN-BERT uses an attention-based relational graph convolutional network with BERT to exploit syntactic information. We will incorporate our aspect-specific context modeling methods into PLMs to see whether the proposed methods can further improve the OE performance.

\section{Aspect-specific Context Modeling}

\begin{figure*}[ht]
  \centering
  \includegraphics[width=\linewidth]{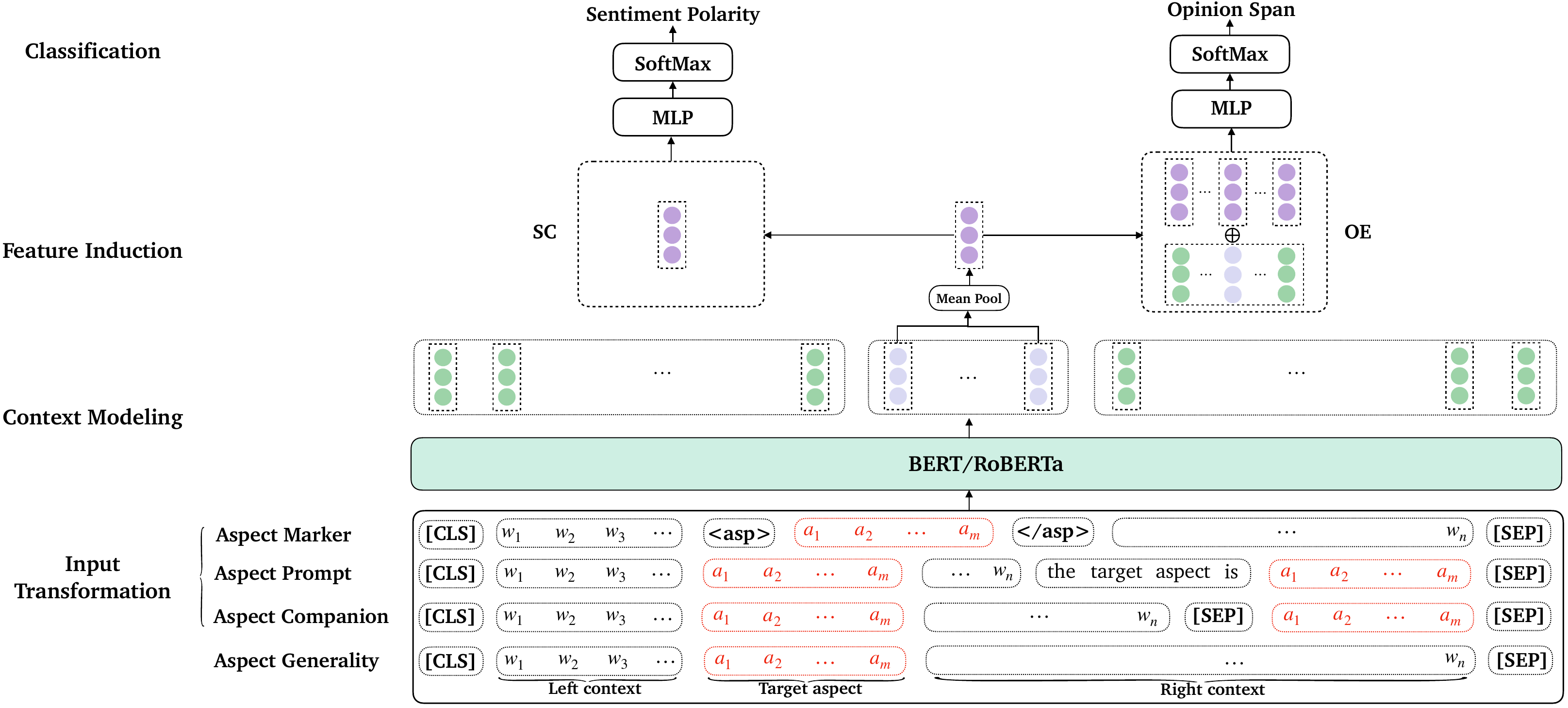}
  \caption{The architecture of our proposed model based on the three mechanisms.}
  \label{Model_Architecture}  
\end{figure*}

\subsection{Task Description}

ABSA (Both SC and OE) requires a pre-given aspect. Formally, a sentence is depicted as $S = \{{w_1, w_2,\dots, w_n}\}$  that contains $n$ words including the aspect. The aspect $A = \{a_1, a_2, ..., a_m\}$ is composed of $m$ words. The goal of SC is to find the sentiment polarity with respect to the given aspect $A$. OE aims to extract corresponding opinion span based on the given aspect $A$. Recap the example in Figure~\ref{absa_example} that contains aspect \textit{food}. SC requires a model to give a \texttt{positive} sentiment on \textit{food} and OE requires a model to tag the sentence as \{\texttt{O}, \texttt{O}, \texttt{O}, \texttt{B}, \texttt{O}, \texttt{O}, \texttt{O}, \texttt{O}, \texttt{O}, \texttt{O}, \texttt{O},\}, indicating the opinion span \textit{tasty} for the aspect \textit{food}. 

\subsection{Overall Framework}
Figure~\ref{Model_Architecture} shows the structure of our model. Conventionally, an ABSA model consists of four parts: an input layer, a context modeling layer, a feature induction layer, and a classification layer. For aspect-specific context modeling, we first use an aspect-specific transformation to enrich the input. Next, the PLM is applied to get contextualized representations. Then we apply a mean pool operation on the hidden states of the first and last aspect tokens to induct the aspect-specific feature. For SC, we use the aspect-specific feature as the final representation for sentiment classification. For OE, we concatenate the aspect-specific feature and each token's representation to form the final representation for span tagging.

\subsection{Aspect-general Input}

The PLM requires a special classification token \texttt{[CLS]} (BERT) or \texttt{$\langle$s$\rangle$} (RoBERTa) be appended to the start of the input sequence, and a separation token \texttt{[SEP]} (BERT) or \texttt{$\langle$/s$\rangle$} (RoBERTa) appended to the end of the input sequence. The original input sentence is converted to the format \texttt{[CLS]} + input sequence + \texttt{[SEP]}. We refer to this format as aspect-general input, termed as \textbf{aspect generality}. Most previous work uses it for ABSA tasks, and \texttt{[CLS]} is often used for downstream classification, but there is no clear aspect information and no way of knowing which aspect is the focus.

\subsection{Aspect-specific Input Transformations}

We propose three aspect-specific input transformations at the input layer to highlight the aspect in the sentence, namely aspect companion, aspect prompt, and aspect marker. We hypothesize that the three transformations can promote the aspect-awareness of PLM and help PLM achieve an effective aspect-specific context modeling.

\subsubsection{Aspect Companion}

Inspired by BERT's sentence pair encoding fashion, previous work \citep{xu2019bert} appends the aspect to the sentence as auxiliary information. Let $\hat{S}$ denote the modified sequence with aspect companion: $\hat{S}=\{\texttt{[CLS]},w_1,\dots,a_1, \dots,a_m,\dots,w_n,\texttt{[SEP]}, a_1,\\ \dots,a_m, \texttt{[SEP]}\}$. This formatted sequence can help the PLM effectively model the intra-sentence dependencies between every pair of tokens and further enhance the inter-sentence dependencies between the global context and the aspect.

\subsubsection{Aspect Prompt} 

Inspired by recently popular prompt tuning where some natural language prompts can make the PLM complete a task in a cloze-completion style~\citep{brown2020language, schick2021s}, we here append to the sentence with an aspect-oriented prompt sentence. Let $\hat{S}$ denote the modified sequence with aspect prompt: $\hat{S}=\{\texttt{[CLS]},w_1,\dots,a_1,\dots,a_m,\dots,w_n,\textrm{the},\textrm{target},\\ \textrm{aspect},\textrm{is},a_1,\dots,a_m,\texttt{[SEP]}\}$. This format sequence prompts the PLM to target at the aimed aspect.

\subsubsection{Aspect Marker}

Aspect marker inserts markers into the sentence to explicitly mark the boundaries of the concerned aspect. Specifically, we define the markers as two preserved tokens: \texttt{$\langle$asp$\rangle$} and \texttt{$\langle$/asp$\rangle$}. We insert them into the input sentence before and after the concerned aspect, to mark the start and end of the given aspect. \texttt{$\langle$asp$\rangle$} indicates the start of the aspect, and \texttt{$\langle$/asp$\rangle$} indicates the end of the aspect. Let $\hat{S}$ denote the modified sequence with aspect marker inserted:
$\hat{S}$ = $\{\texttt{[CLS]}, w_1, \dots , \texttt{$\langle$asp$\rangle$}, a_1, \dots, a_m, \texttt{$\langle$/asp$\rangle$}, \dots ,\\ w_n, \texttt{[SEP]}\}$.

The three \textit{aspect-specific input transformations} gain significant improvement in our experiments (Section~\ref{section:results}), and this strengthens our hypothesis that injecting the aspect marker at the input layer can help the PLM capture aspect-specific contextual information further.


\subsection{Context Modeling}

Previous PLM-based ABSA work directly adopts the hidden states of the PLM for downstream classification. However, an empirical observation is that the context words close to the aspect are more semantic-relevant to the aspect \citep{ma2021exploiting}. In the case, more sentiment information is possibly contained in the aspect's local context rather than the global context. As a result, the general usage of the hidden states from the PLM loses much local contextual information related to the aspect. With the help of the three input transformations, we obtain the hidden states that incorporate the aspect-oriented local context. Let
\begin{equation}
    H = \mathrm{PLM}(\hat{S})
    \label{embedding1}
\end{equation}
where $H = {\{h_1,h_2, \dots ,h_1^a, \dots , h_m^a, \dots , h_n\}}$ represents the sequence of hidden states.


\subsection{Feature Induction}
As aforementioned, aspect-general feature induction contains the semantic information critical to the whole sentence rather than the given aspect, and the induced aspect-general feature may be aspect-irrelevant when the sentence contains two or more aspects. After getting the global contextual representation $H$, existing work needs an aspect-specific feature extraction strategy to induce the aspect feature after getting the global contextual representation $H$. For an enriched aspect-awareness, we adopt the mean pool on the hidden states corresponding to the first and last aspect tokens. Let
\begin{equation}
    \hat{H} = \mathrm{MeanPool}([h_1^a, h_m^a])
    \label{embedding2}
\end{equation}
represent the aspect-specific feature, where $h_1^a$ indicates the hidden state of the first aspect token, and $h_m^a$ indicates the hidden state of the last aspect token. Due to that OE is a token-level classification task, we concatenate the aspect-specific feature $\hat{H}$ and the global contextual representation $H$ as the final aspect-specific contextual representation for tagging. 

\subsection{Fine-tuning}
After getting the aspect-specific contextual representation $\hat{H}$, an multi-layered Perceptron (MLP) layer is used to fine-tune the proposed BERT or RoBERTa based model. The MLP contains four steps: a fully-connected layer, a ReLU activation function layer, a dropout layer, and a fully-connected layer. Then we feed the output to a softmax layer to predict the corresponding label. The training objective is to minimize the cross-entropy loss with $\mathcal{L}_{2}$ regularization. Specifically, the optimal parameters $\theta$ are obtained from 

\begin{equation}
    \mathcal{L}(\theta) = - \sum_{i=1}^{n} \hat{y}_i \log{y_i} + \lambda \sum_{\theta \in \Theta}^{} {\theta}^2
\end{equation}
where $\lambda$ is the regularization constant and $\hat{y_i}$ is the predicted label corresponding to ground truth label $y_i$.

When no input transformation is used, the model is aspect-general and named as PLM-MeanPool and PLM-MeanPool-Concat for SC and OE, respectively. By incorporating the three input transformations, the model becomes more aspect-specific, denoted as +AC (Aspect Companion), +AP (Aspect Prompt), and +AM (Aspect Marker) respectively.


\section{Experiments}

\subsection{Datasets}

\subsubsection{SC Datasets} 

Following previous work \citep{ma2021exploiting}, we conduct experiments on two SC benchmarks to evaluate our models' effectiveness and robustness. One is SemEval 2014 \citep{pontikisemeval} (\textsc{SemEval}), which contains data from laptop (\textsc{Sem-Lap}) and restaurant (\textsc{Sem-Rest}) domains; the other is the Aspect Robustness Test Set (\textsc{ARTS-SC})~\citep{xing2020tasty}, which is derived from the \textsc{SemEval} dataset. Instances in \textsc{ARTS-SC} are generated with three adversarial strategies. These strategies enrich the test set from 638 to 1,877 for the laptop domain (\textsc{Arts-SC-Lap}), and from 1,120 to 3,530 for the restaurant domain (\textsc{ARTS-SC-Rest}). Note that each domain from \textsc{SemEval} consists of separate training and test sets, while each domain from \textsc{ARTS-SC} only contains a test set. Since datasets in \textsc{SemEval} do not contain development sets, 150 instances from the training set in each dataset are randomly selected to form the development set. Table \ref{ascdatasets} shows the statistics of the SC datasets.

\subsubsection{OE Datasets} 

For datasets used in OE \citep{fan2019target, wu2020latent}, the original \textsc{SemEval} benchmark annotates the aspects, but not the corresponding opinion spans, for each sentence. To solve the problem, \citep{fan2019target} annotates the corresponding opinion spans for each given aspect in a sentence and removes the cases without explicit opinion spans. We use this variant in our OE experiments.

Since there is currently no robustness test set for OE, we follow ~\citep{xing2020tasty}'s three adversarial strategies to generate an Aspect Robustness Test Set with spans (\textbf{\textsc{ARTS-OE}}) based on \textsc{SemEval}. Specifically, we use these strategies to generate 1002 test instances for the laptop domain (\textbf{\textsc{ARTS-OE-Lap}}) and 2009 test instances for the restaurant domain (\textbf{\textsc{ARTS-OE-Res}}). Each aspect in a sentence is associated with an opinion span for OE. It is worth noting that this adversarial dataset can also be used for other tasks, e.g., aspect sentiment triplet extraction ~\citep{peng2020knowing}. Table \ref{AOEdatasets} shows the statistics of the OE datasets. And the details of \textsc{ARTS-OE} are provided in Tabel~\ref{ARTS_Span_Example}. Since these OE datasets do not come with a development set, we randomly split 20\% of the training set as validation set.

 \begin{table}[ht]
\centering
\resizebox{0.40\textwidth}{!}{
\begin{tabular}{ccccc}
\hline
\multicolumn{2}{c}{\textbf{Dataset}}       & \textbf{\#pos.} & \textbf{\#neu.} & \textbf{\#neg.} \\ 
\hline
\multirow{3}{*}{\textsc{\textbf{Sem-Lap}}}     & train & 930      & 433     & 800      \\
\cline{2-5}
                            & test  & 341      & 169     & 128      \\
                            \cline{2-5}
                            & dev   & 57       & 27      & 66       \\ 
                            \hline
\multirow{3}{*}{\textsc{\textbf{Sem-Rest}}} & train & 2,094     & 579     & 779      \\
\cline{2-5}
                            & test  & 728      & 196     & 196      \\
\cline{2-5}
                            & dev   & 70       & 54      & 26       \\
\hline
\textsc{\textbf{Arts-SC-Lap}}                    & test  & 883      & 407     & 587      \\ 
\hline
\textsc{\textbf{Arts-SC-Rest}}                  & test  & 1,953     & 473     & 1,104     \\ 
\hline
\end{tabular}}
\caption{Statistics of SC datasets.}
\label{ascdatasets}  
\end{table}

\begin{table}[ht]
\centering
\resizebox{0.45\textwidth}{!}{
\begin{tabular}{clccc}
\hline
\multicolumn{3}{c}{\textbf{Dataset}}                                 & \textbf{\#sentences} & \textbf{\#aspects} \\ \hline
\multicolumn{2}{c}{\multirow{2}{*}{\textsc{\textbf{Sem-Lap}}}} & train        & 1,158        & 1,634      \\ \cline{3-5}
\multicolumn{2}{c}{}                                & test         & 343         & 482       \\ \hline
\multicolumn{2}{c}{\multirow{2}{*}{\textsc{\textbf{Sem-Rest}}}} & train        & 1,627        & 2,643      \\ \cline{3-5}
\multicolumn{2}{c}{}                                & test         & 500         & 865       \\ \hline

\multicolumn{2}{c}{\multirow{1}{*}{\textsc{\textbf{ARTS-OE-Lap}}}} & test        & 1,002         & 2,404         \\ \hline
\multicolumn{2}{c}{\multirow{1}{*}{\textsc{\textbf{ARTS-OE-Rest}}}} & test        & 2,009        & 5,743      \\ \hline
\end{tabular}}
\caption{Statistics of OE datasets. A sentence may contain multiple aspects. The number of aspect is identical to the number of pairs and instances.}
\label{AOEdatasets} 
\end{table}

\subsection{Comparative Models and Baselines}
We carry out an extensive evaluation of the proposed models (with and without input transformation), including \textbf{\textit{PLM-MeanPool}} and \textbf{\textit{PLM- MeanPool +AC/AP/AM}} for SC, \textbf{\textit{PLM-MeanPool-Concat}} and \textbf{\textit{PLM-MeanPool-Concat +AC/AP/AM}} for OE, against a wide range of baselines, categorized into two groups: PLM-based models and non-PLM models.  


\subsubsection{SC Baselines}

\textbf{\textit{Non-PLM models}} include: (a) \underline{IAN} \citep{ma2017interactive} interactively learns attentions between context words and aspect terms. (b) \underline{MemNet} \citep{tang2016aspect} applies attention multiple times on word memories, and the output of the last attention is used for prediction. (c) \underline{AOA} \citep{huang2018aspect} introduces an attention-over-attention based network to model interaction between aspects and contexts. (d) \underline{ASGCN} \citep{zhang2019aspect} use graph convolutional networks to capture the aspect-specific information. \textbf{\textit{PLM-based models}} include: (a)  \underline{BERT/RoBERTa-CLS-MLP} use the representation of "\texttt{[CLS]}" as a classification feature to fine-tune the BERT/RoBERTa model with an MLP layer. (b) \underline{AEN-BERT} \citep{song2019attentional} adopts BERT model and attention mechanism to model the relationship between contexts and aspects. (c) \underline{LCF-BERT} \citep{zeng2019lcf} employs Local-Context-Focus design with Semantic-Relative-Distance to discard unrelated sentiment words. (d) \underline{BERT/RoBERTa-ASCNN} is combined with BERT/RoBERTa and ASCNN \citep{zhang2019aspect} model. (e)\underline{RoBERTa-ASGCN}~\cite{zhang2019aspect} is combined with RoBERTa and ASGCN.



\subsubsection{OE Baselines}

\textbf{\textit{Non-PLM models}} include:
(a) \underline{Pipeline} \citep{fan2019target} is a combination method of BiLSTM and Distance-rule method \citep{hu2004mining}. (b) \underline{IOG} \citep{fan2019target} utilizes an Inward-Outward LSTM and a Global LSTM to capture the information of aspect and global information, respectively. (c) \underline{LOTN} Latent Opinions Transfer Network \citep{wu2020latent} uses an effective transfer learning method to identify latent opinions from the sentiment analysis model. (d) \underline{ARGCN} \citep{jiang2021attention} is an extension of R-GCNs suited to encode syntactic dependency information to complete OE. \textbf{\textit{PLM-based models}} include: (a) \underline{BERT+Distance-rule} \citep{feng2021target} is the combination of BERT and Distance-rule. (b) \underline{TF-BERT} \citep{feng2021target} utilizes the average pooling of target word embeddings to represent the target information, then it is fed into BERT to extract target-oriented opinion terms. (c) \underline{SDRN} \citep{chen2020synchronous} utilizes BERT as the encoder, which consists of an opinion entity extraction unit, a relation detection unit, and a synchronization unit for the aspect opinion pair extraction task. (d) \underline{TSMSA-BERT} \citep{feng2021target} uses a target-specified sequence labeling method based on multi-head self-attention (TSMSA) to perform OE. (e) \underline{ARGCN+BERT} \citep{jiang2021attention} adopts the last hidden states of the pretrained BERT as word representations and fine-tune it with the ARGCN model.


Implementation details and evaluation metrics can be found in Appendix \ref{section:Implementation_details} and \ref{section:Evaluation_metrics}. It is worth noting that most previous methods did not use the dev set and may have overfitted the test set. We have made a systematic and comprehensive comparison for the first time under the same settings.


\section{Results and Analysis} \label{section:results}
\subsection{SC Results}
Table~\ref{asc_result} shows the standard (effectiveness) and robustness evaluation results for SC. 

\subsubsection{Standard Results}


 Generally, our models with input transformations outperform the comparative baseline models. Before applying the transformations, our base models (BERT/RoBERTa-MeanPool with aspect generality) perform equally good or even better than most baseline models. 
 
 Applying the input transformations, especially aspect marker (i.e., +AM), further improves performance significantly. For BERT-based models, the F1-scores of the BERT-MeanPool+AM model are 2.57\% and 5.83\% higher than AEN-BERT and LCF-BERT respectively on the \textsc{Sem-Rest} standard dataset. For RoBERTa-based models, the three transformations are more effective. Specifically, the F1-scores of RoBERTa-MeanPool+AC and RoBERTa-MeanPool+AP improve by up to 1.54\% and 1.28\% on \textsc{Sem-Rest} standard dataset. These results indicate that the proposed input transformations can promote PLMs to achieve effective aspect-specific context modeling. 
 
 Among the three transformations, in general AM performs better than AC and AP, indicating that AM is more effective for  aspect-specific context modeling in PLMs. While the F1-scores of BERT-MeanPool+AM and RoBERTa-MeanPool+AM gain improvements by 1.59\% and 1.43\% on \textsc{Sem-Rest},  RoBERTa-MeanPool+AM achieves the terrific results for SC, with F1-score are 78.5\% and 79.58\% on \textsc{Sem-Lap} and \textsc{Sem-Rest} respectively.


\begin{table*}[ht]
\centering
 \resizebox{0.86\textwidth}{!}{
\begin{tabular}{c|cccc|cccc}
\hline
\multirow{3}{*}{\textbf{Models}}     & \multicolumn{4}{c|}{\textbf{\textsc{Sem-Lap}}}                                                & \multicolumn{4}{c}{\textbf{\textsc{Sem-Rest}}}                                                                        \\ \cline{2-9} 
                                     & \multicolumn{2}{c}{\textbf{Standard}} &  \multicolumn{2}{c|}{\textbf{Robustness}} & \multicolumn{2}{c}{\textbf{Standard}} &  \multicolumn{2}{c}{\textbf{Robustness}} \\ \cline{2-9} 
                                     & \textbf{Acc.}    & \textbf{F1}    &  \textbf{Acc.}     & \textbf{F1}    & \textbf{Acc.}    & \textbf{F1}    &  \textbf{Acc.}    & \textbf{F1}    \\ \hline
IAN                                  & 67.74            & 59.99              & 52.91             & 47.54          & 77.48            & 66.39          & 57.75            & 48.12          \\
Memnet                               & 67.81            & 60.67            & 52.00             & 46.50          & 76.77            & 64.46            & 55.30            & 46.67          \\
AOA                                  & 69.47            & 63.13            & 52.00             & 46.50          & 77.57            & 66.02           & 58.19            & 49.02          \\
ASGCN    & 70.97            & 65.31               & 56.59             & 52.12          & 78.87            & 68.12           & 64.89            & 55.41          \\ 

\hline
AEN-BERT                             & 77.37            & 71.83                    & 71.49             & 66.37          & 83.66            & 75.50                  & 73.24            & 66.31          \\
LCF-BERT                             & 76.55            & 71.40                   & 71.19             & 66.95          & 81.66            & 72.24                & 70.57            & 62.75          \\
BERT-CLS+MLP                         & 75.42            & 69.08           & 54.91             & 51.21          & 78.95            & 67.66                 & 53.86            & 47.16          \\
RoBERTa-CLS+MLP                      & 79.09            & 75.36                   & 56.24             & 54.61          & 81.93            & 71.19                  & 60.45            & 52.02          \\
BERT-ASCNN                           & 76.33            & 71.09                 & 71.17             & 66.90          & 82.66            & 74.05              & 75.73            & 68.17          \\
RoBERTa-ASCNN                        & 81.41            & 77.22                   & 73.59             & 70.14          & 85.93            & 78.01                   & 78.85            & 70.69          \\
RoBERTa-ASGCN  & 81.82            & 78.28        & 73.48             & 69.38          & 85.66            & 78.48                   & 79.65            & 72.56          \\
\hline
\textbf{BERT-MeanPool}                            & 76.87            & 71.71                  & 70.59             & 66.38          & 84.27            & 76.48                  & 77.36            & 70.64          \\
+\textbf{AC}   & 75.30            & 69.62                   & 69.40             & 64.45          & 84.12            & 76.16               & 76.78            & 69.86          \\

+\textbf{AP}   & 76.39            & 70.91                 & 68.92             & 63.77          & 83.89            & 76.02       & 76.48            & 69.34          \\ 
+\textbf{AM}    & 76.33            & \textbf{71.93}\small{$\uparrow$\textbf{{\textcolor{red}{0.22}}}}                  & \textbf{70.78}             & \textbf{67.06}\small{$\uparrow$\textbf{{\textcolor{red}{0.68}}}}          & \textbf{84.71}            & \textbf{78.07} \small{$\uparrow$\textbf{{\textcolor{red}{1.59}}}}                 & \textbf{78.10}            & \textbf{72.38} \small{$\uparrow$\textbf{{\textcolor{red}{1.74}}}}         \\

\hline
\textbf{RoBERTa-MeanPool}                          & 81.38            & 77.68                    & 74.67             & 71.21          & 85.41            & 78.15                 & 79.75            & 72.73          \\

+\textbf{AC} & \textbf{81.54}            & 77.54                    & \textbf{75.13}             & 71.02          & \textbf{86.68}\textcolor{red}{$^{\dagger}$}            & \textbf{79.69}\textcolor{red}{$^{\dagger}$}\small{$\uparrow$\textbf{{\textcolor{red}{1.54}}}}                   & \textbf{80.63}            & \textbf{74.03}\small{$\uparrow$\textbf{{\textcolor{red}{1.30}}}}           \\

+\textbf{AP} & \textbf{81.85}            & \textbf{77.91}\small{$\uparrow$\textbf{{\textcolor{red}{0.23}}}} 
   & 74.53            & 70.48
& \textbf{86.43}            & \textbf{79.43}\small{$\uparrow$\textbf{{\textcolor{red}{1.28}}}} 
& \textbf{80.72}             & \textbf{74.09}\textcolor{red}{$^{\dagger}$}\small{$\uparrow$\textbf{{\textcolor{red}{1.36}}}}             \\
+\textbf{AM}  & \textbf{82.07}\textcolor{red}{$^{\dagger}$}            & \textbf{78.50}\textcolor{red}{$^{\dagger}$}\small{$\uparrow$\textbf{{\textcolor{red}{0.82}}}}                  & \textbf{75.90}\textcolor{red}{$^{\dagger}$}             & \textbf{72.59}\textcolor{red}{$^{\dagger}$}\small{$\uparrow$\textbf{{\textcolor{red}{1.38}}}}          & \textbf{86.41}            & \textbf{79.58}\small{$\uparrow$\textbf{{\textcolor{red}{1.43}}}}         & \textbf{80.88}\textcolor{red}{$^{\dagger}$}            & \textbf{74.04}\small{$\uparrow$\textbf{{\textcolor{red}{1.31}}}}          \\

\hline
\end{tabular}
}
\caption{Standard and robust experimental results (\%) on SC. The first and second blocks indicate non-PLM and PLM-based baseline models. Our models and better results are bold (Acc and F1, the larger, the better). The marker \textcolor{red}{$^{\dagger}$} represents that our models outperform the all other models significantly (p ${<}$ 0.01), and the small number next to each score indicates performance improvement ($\uparrow$) compared with our aspect-general base model (BERT-MeanPool/RoBERTa-MeanPool).}
\label{asc_result}  
\end{table*}


\subsubsection{Robustness Results}
 We can see that the performances of the baseline models drop drastically on robustness test sets. In contrast, our models with the three transformations are more robust than the baseline models. The most effective and robust model is the RoBERTa-MeanPool+AM, which achieves 72.59\% and 74.04\% of F1 score on the \textsc{ARTS-SC-LAP} and \textsc{ARTS-SC-REST} robustness test set, respectively, representing a 3.21\% and 1.48\% improvement over the strongest baseline RoBERTa-ASGCN.
 
 The three transformations significantly improve the BERT/RoBERTa-MeanPool models' robustness, especially for RoBERTa-MeanPool. Specifically, with AC, AP, and AM, the RoBERTa-MeanPool model's F1-scores are improved by up to 1.30\%, 1.36\%, and 1.31\% on \textsc{ARTS-SC-Rest} robustness test set, respectively. The model with AM is more robust than the model with AC and AP. These robustness results demonstrate that the transformations can improve our models' robustness.

\begin{table*}[ht]
\centering
\resizebox{0.62\textwidth}{!}{
\begin{tabular}{c|cc|cc}
\hline
\multirow{2}{*}{\textbf{Models}} & \multicolumn{2}{c|}{\textbf{\textsc{Sem-Lap}}}                          & \multicolumn{2}{c}{\textbf{\textsc{Sem-Rest}}}   

\\ \cline{2-5}
 & \textbf{Standard}    & \textbf{Robustness}   & \textbf{Standard}   & \textbf{Robustness}                                                                                  \\ \hline
Pipeline*                                         & 63.83     & -     & 69.18     & -   \\
IOG*                                              & 70.99     & -           & 80.23   & -               \\
LOTN*                                             & 72.02      & -    & 82.21   & -    \\
ARGCN*                                            & 75.32      & -    & 84.65   & -    \\ \hline
BERT+Distance-rule*                               & 70.54     & -         & 76.23  & -          \\
TF-BERT*                                          & 72.26      & -     & 78.23     & -     \\
SDRN*                                             & 80.24    & -    & 83.53      & -          \\
TSMSA-BERT*                                        & 82.18     & -    & 86.37    & -         \\
ARGCN-BERT*                                       & 76.36                            & -        & 85.42               & -                                          \\ \hline
\textbf{BERT-MeanPool-Concat}              & 68.27                                   & 39.68            & 69.08      & 44.23                                        \\
\textbf{+AC}             & 80.31\small{$\uparrow$\textbf{{\textcolor{red}{12.04}}}}                                                                           & 70.98\small{$\uparrow$\textbf{{\textcolor{red}{31.30}}}}                                                                           & 85.09\small{$\uparrow$\textbf{{\textcolor{red}{16.01}}}}                                                                          & 70.01\small{$\uparrow$\textbf{{\textcolor{red}{25.78}}}}                                                                           \\
\textbf{+AP}              & 79.60\small{$\uparrow$\textbf{{\textcolor{red}{11.33}}}}                                                                           & 68.06\small{$\uparrow$\textbf{{\textcolor{red}{28.38}}}}                                                                           & 85.32\small{$\uparrow$\textbf{{\textcolor{red}{16.24}}}}                                                                          & 70.25\small{$\uparrow$\textbf{{\textcolor{red}{26.02}}}}                                                                           \\
\textbf{+AM}              & 81.06 \small{$\uparrow$\textbf{{\textcolor{red}{12.79}}}}                                                                          & 71.23\small{$\uparrow$\textbf{{\textcolor{red}{31.55}}}}                                                                           & 85.62\small{$\uparrow$\textbf{{\textcolor{red}{16.54}}}}                                                                          & 69.68\small{$\uparrow$\textbf{{\textcolor{red}{25.45}}}}                                                                           \\ \hline
\textbf{RoBERTa-MeanPool-Concat}            & 69.74                                                                  & 38.76                                                                & 79.03      & 56.93                                     \\
\textbf{+AC}           & 82.78\small{$\uparrow$\textbf{{\textcolor{red}{13.04}}}}                                                                           & 71.26\small{$\uparrow$\textbf{{\textcolor{red}{32.50}}}}                                                                           & 86.03\small{$\uparrow$\textbf{{\textcolor{red}{7.00}}}}                                                                           & 71.42\small{$\uparrow$\textbf{{\textcolor{red}{14.49}}}}                                                                           \\
\textbf{+AP}            & 82.63\small{$\uparrow$\textbf{{\textcolor{red}{12.89}}}}                                                                           & 71.46\small{$\uparrow$\textbf{{\textcolor{red}{32.30}}}}                                                                           & \textbf{86.58}\textcolor{red}{$^{\dagger}$}\small{$\uparrow$\textbf{{\textcolor{red}{7.55}}}} & \textbf{71.61}\textcolor{red}{$^{\dagger}$}\small{$\uparrow$\textbf{{\textcolor{red}{14.68}}}} \\
\textbf{+AM}            & \textbf{83.83}\textcolor{red}{$^{\dagger}$}\small{$\uparrow$\textbf{{\textcolor{red}{14.09}}}} & \textbf{73.69}\textcolor{red}{$^{\dagger}$}\small{$\uparrow$\textbf{{\textcolor{red}{34.93}}}} & 86.33\small{$\uparrow$\textbf{{\textcolor{red}{7.30}}}}                                                                           & 71.50\small{$\uparrow$\textbf{{\textcolor{red}{14.57}}} }                                                   \\ \hline
\end{tabular}}
\caption{Standard and robustness evaluation results (F1-score, \%) on OE. The first and second blocks show the results of the non-PLM and BERT-based baseline models (with $*$) respectively, which are extracted from the published papers \citep{wu2020latent} and \citep{feng2021target}. Note that there were no robustness results of the baseline models in the original published papers, so that we leave then blank. The results of our models are presented in the third and fourth blocks. The best results are bold (F1-score, the larger, the better). }.
\label{AOE_result}  
\end{table*}

\subsection{OE Results}
Tabel~\ref{AOE_result} shows the standard and robustness results for OE.

\subsubsection{Standard Results}
 Before applying the transformations, our base models (BERT/RoBERTa-MeanPool-Concat) perform poorly, even worse than most non-PLM baseline models. On the contrary, with the three transformations, our models perform significantly better than baseline models. Our BERT-based model with the three transformations achieves nearly identical results with the current state-of-the-art model (TSMSA-BERT). With AC, AP, and AM, the F1-scores of the RoBERTa-MeanPool-Concat model are improved by up to 13.04\%, 12.89\%, and 14.09\% on \textsc{Sem-Lap} dataset, respectively. These results demonstrate that the three transformations can significantly promote PLMs to achieve effective aspect-specific context modeling for OE. Our RoBERTa-MeanPool-Concat+AM model achieves the new state-of-the-art result on OE.

\subsubsection{Robustness Results}
The performances of our base models (BERT /RoBERTa-MeanPool-Concat) drop drastically on robustness test set. Their F1-scores are only 39.68\% and 38.76\% on \textsc{ARTS-OE-Lap} and 44.23\% and 56.93\% on \textsc{ARTS-OE-Rest}. In contrast, with the transformations, our models are more robust, achieving F1 scores up to 73.69\% (RoBERTa-MeanPool-Concat+AM) on \textsc{ARTS-OE-Lap}, and 71.61\% (RoBERTa-MeanPool-Concat+AP) on \textsc{ARTS-OE-Rest}, demonstrating that the transformations can significantly improve our model's robustness for OE.

\begin{table}[ht]
\centering
\resizebox{0.38\textwidth}{!}{
\begin{tabular}{c|cc}
\hline
\textbf{Models} & SEM-LAP & SEM-REST \\ \hline
BERT-MeanPool    & 71.71   & 76.48    \\
\hline
BERT-CLS+MLP    & 69.08   & 67.66    \\
+AC             & 68.82   & 74.03\small{$\uparrow$\textbf{{\textcolor{red}{6.37}}}}    \\
+AP             & 70.47\small{$\uparrow$\textbf{{\textcolor{red}{1.39}}}}   & 76.78\small{$\uparrow$\textbf{{\textcolor{red}{9.12}}}}    \\
+AM             & 70.24\small{$\uparrow$\textbf{{\textcolor{red}{1.16}}}}   & 74.19\small{$\uparrow$\textbf{{\textcolor{red}{6.53}}}}    \\ \hline
RoBERTa-MeanPool    & 77.68   & 78.15    \\ 
\hline
RoBERTa-CLS+MLP & 75.36   & 71.19    \\
+AC             & 77.62\small{$\uparrow$\textbf{{\textcolor{red}{2.26}}}}   & 76.04\small{$\uparrow$\textbf{{\textcolor{red}{4.85}}}}    \\
+AP             & 78.40\small{$\uparrow$\textbf{{\textcolor{red}{3.04}}}}   & 78.53\small{$\uparrow$\textbf{{\textcolor{red}{7.34}}}}    \\
+AM             & 78.21\small{$\uparrow$\textbf{{\textcolor{red}{2.85}}}}   & 79.91\small{$\uparrow$\textbf{{\textcolor{red}{8.72}}}}    \\ \hline
\end{tabular}}
\caption{SC ablation experimental results (F1-score, \%). }
\label{SC_Ablation_result}  
\end{table}

\begin{table}[ht!]
\centering
\resizebox{0.40\textwidth}{!}{
\begin{tabular}{c|cc}
\hline
\textbf{Models} & SEM-LAP & SEM-REST \\ \hline
BERT-MeanPool-Concat        & 68.27   & 69.08    \\
\hline
BERT-MLP        & 67.67   & 61.40    \\
+AC             & 79.95\small{$\uparrow$\textbf{{\textcolor{red}{12.28}}}}   & 79.46\small{$\uparrow$\textbf{{\textcolor{red}{18.06}}}}    \\
+AP             & 80.08\small{$\uparrow$\textbf{{\textcolor{red}{12.41}}}}   & 81.02\small{$\uparrow$\textbf{{\textcolor{red}{19.62}}}}    \\
+AM             & 81.50\small{$\uparrow$\textbf{{\textcolor{red}{13.83}}}}   & 80.02\small{$\uparrow$\textbf{{\textcolor{red}{18.62}}}}    \\ \hline
RoBERTa-MeanPool-Concat        & 69.74   & 79.03    \\
\hline
RoBERTa-MLP     & 67.92   & 60.00    \\
+AC             & 82.18\small{$\uparrow$\textbf{{\textcolor{red}{14.26}}}}   & 81.59\small{$\uparrow$\textbf{{\textcolor{red}{21.59}}}}    \\
+AP             & 81.96\small{$\uparrow$\textbf{{\textcolor{red}{14.04}}}}   & 81.04 \small{$\uparrow$\textbf{{\textcolor{red}{21.04}}}}   \\
+AM             & 83.42\small{$\uparrow$\textbf{{\textcolor{red}{15.50}}}}   & 80.81\small{$\uparrow$\textbf{{\textcolor{red}{20.81}}}}    \\ \hline
\end{tabular}}
\caption{OE ablation experimental results (F1-score). }
\label{OE_Ablation_result} 
\end{table}

\subsection{Ablation Study}
To further investigate the effects of the feature induction and the input transformations on aspect-specific context modeling of PLMs, we conduct extensive ablation experiments on standard datasets, whose results are included in Table~\ref{SC_Ablation_result} and~\ref{OE_Ablation_result} for SC and OE, respectively.

\subsubsection{Aspect-specific Feature Induction} For SC and OE, we start with a simple base model that does not use the aspect feature induction component, but using just a context modeling representation after PLM and append an MLP layer. The base model is named as BERT/RoBERTa-CLS-MLP for SC, and BERT/RoBERTa-MLP for OE. Now we see what happens if we add back the aspect feature induction. For SC, our BERT/RoBERTa-MeanPool models always give a superior performance than the base model. The F1-scores of BERT-MeanPool are 2.63\% and 8.82\% higher than BERT-CLS-MLP on \textsc{Sem-Lap} and \textsc{Sem-REST} respectively. For OE, our BERT/RoBERTa-MeanPool-Concat models perform better than BERT/RoBERTa-MLP models. These results demonstrate the effectiveness of the aspect-specific feature induction methods with PLMs.

\subsubsection{Aspect-specific Context Modeling} To investigate the effect of the aspect-specific context modeling with transformations, we add the input transformations to the above simple base models. The results show that the transformations bring significant performance improvements, even better than the models with aspect feature induction. Especially the base models with the transformations for OE achieve nearly identical results to BERT/RoBERTa-MeanPool-Concat with transformations. These excellent results demonstrate the effectiveness of the proposed transformations for context modeling, which indirectly explains that context modeling is more critical than aspect feature induction for ABSA.

\begin{figure}[ht]
\centering
\resizebox{0.48\textwidth}{!}{
\begin{tabular}{cc}
\hline  
\textbf{Model} & \textbf{Example}     \\  
\hline
AG & 
\colorbox{orange!100.0}{\strut [CLS]} \colorbox{orange!28.76}{\strut The} \colorbox{orange!28.05}{\strut \underline{food}} \colorbox{orange!30.30}{\strut is} \colorbox{orange!54.41}{\strut \underline{ta}} \colorbox{orange!40.35}{\strut \underline{\#\#sty}} \colorbox{orange!35.27}{\strut but} \colorbox{orange!29.64}{\strut the} \colorbox{orange!26.19}{\strut service} \colorbox{orange!32.79}{\strut is} \colorbox{orange!82.65}{\strut bad} \colorbox{orange!44.73}{\strut !} \colorbox{orange!58.22}{\strut [SEP]} 
\\ 
\cline{2-2}
 AC & 
\colorbox{orange!100.0}{\strut [CLS]} \colorbox{orange! 45.92}{\strut The} \colorbox{orange! 47.25}{\strut \underline{food}} \colorbox{orange!50.45}{\strut is} \colorbox{orange!72.29}{\strut \underline{ta}} \colorbox{orange!78.30}{\strut \underline{\#\#sty}} \colorbox{orange! 69.72}{\strut but} \colorbox{orange!52.30}{\strut the} \colorbox{orange!42.84}{\strut service} \colorbox{orange!50.93}{\strut is} \colorbox{orange!52.11}{\strut bad} \colorbox{orange!62.35}{\strut !} \colorbox{orange!60.20}{\strut [SEP]} \colorbox{orange!57.64}{\strut food} \colorbox{orange!60.14}{\strut [SEP]}  
\\
\cline{2-2}
 AP & 
\colorbox{orange!100.0}{\strut [CLS]} \colorbox{orange! 37.84}{\strut The} \colorbox{orange! 36.29}{\strut \underline{food}} \colorbox{orange!43.18}{\strut is} \colorbox{orange!56.02}{\strut \underline{ta}} \colorbox{orange!64.38}{\strut \underline{\#\#sty}} \colorbox{orange! 43.21}{\strut but} \colorbox{orange!42.10}{\strut the} \colorbox{orange!45.60}{\strut service} \colorbox{orange!42.09}{\strut is} \colorbox{orange!54.74}{\strut bad} \colorbox{orange!50.14}{\strut !} \colorbox{orange!44.76}{\strut The} \colorbox{orange!49.51}{\strut target}
\colorbox{orange!64.02}{\strut aspect} \colorbox{orange!45.86}{\strut is}
\colorbox{orange!44.57}{\strut food} \colorbox{orange!31.32}{\strut [SEP]} 
\\
\cline{2-2}

AM & 
\colorbox{orange!100.0}{\strut [CLS]} \colorbox{orange!49.89}{\strut The} \colorbox{orange!33.14}{\strut $<$asp$>$} \colorbox{orange!27.61}{\strut \underline{food}} \colorbox{orange!29.54}{\strut $<$/asp$>$}  \colorbox{orange!57.37}{\strut is} \colorbox{orange!72.69}{\strut \underline{ta}} \colorbox{orange!91.69}{\strut \underline{\#\#sty}} \colorbox{orange!58.87}{\strut but} \colorbox{orange!62.24}{\strut the}  \colorbox{orange!38.13}{\strut service} \colorbox{orange!63.82}{\strut is} \colorbox{orange! 69.54}{\strut bad} \colorbox{orange!74.47}{\strut !} \colorbox{orange! 49.74}{\strut [SEP]} 
\\
\hline
\hline
AG & 
\colorbox{orange!100.0}{\strut [CLS]} \colorbox{orange!28.76}{\strut The} \colorbox{orange!28.05}{\strut food} \colorbox{orange!30.30}{\strut is} \colorbox{orange!54.41}{\strut ta} \colorbox{orange!40.35}{\strut \#\#sty} \colorbox{orange!35.27}{\strut but} \colorbox{orange!29.64}{\strut the} \colorbox{orange!26.19}{\strut \underline{service}} \colorbox{orange!32.79}{\strut is} \colorbox{orange!82.65}{\strut \underline{bad}} \colorbox{orange!44.73}{\strut !} \colorbox{orange!58.22}{\strut [SEP]}  
\\
\cline{2-2}
 AC & 
\colorbox{orange!72.12}{\strut [CLS]} \colorbox{orange!31.94}{\strut The} \colorbox{orange!26.66}{\strut food} \colorbox{orange!33.73}{\strut is} \colorbox{orange!27.58}{\strut ta} \colorbox{orange!24.67}{\strut \#\#sty} \colorbox{orange!38.57}{\strut but} \colorbox{orange!31.37}{\strut the} \colorbox{orange!35.07}{\strut \underline{service}} \colorbox{orange!  29.69}{\strut is} \colorbox{orange!100.0}{\strut \underline{bad}} \colorbox{orange!29.58}{\strut !} \colorbox{orange! 40.18}{\strut [SEP]} \colorbox{orange!38.76}{\strut service} \colorbox{orange!40.10}{\strut [SEP]}  
\\
\cline{2-2}
 AP & 
\colorbox{orange!78.75}{\strut [CLS]} \colorbox{orange!32.53}{\strut The} \colorbox{orange!35.68}{\strut food} \colorbox{orange!37.95}{\strut is} \colorbox{orange!53.85}{\strut ta} \colorbox{orange!50.81}{\strut \#\#sty} \colorbox{orange!62.96}{\strut but} \colorbox{orange!32.98}{\strut the} \colorbox{orange!35.80}{\strut \underline{service}}\colorbox{orange!  34.67}{\strut is} \colorbox{orange!100.0}{\strut \underline{bad}} \colorbox{orange!51.23}{\strut !} \colorbox{orange!38.94}{\strut The } \colorbox{orange!42.82}{\strut target}  \colorbox{orange!53.34}{\strut aspect}  \colorbox{orange!40.22}{\strut is}  \colorbox{orange!45.55}{\strut service}   \colorbox{orange!21.64}{\strut [SEP]}  
\\
\cline{2-2}
AM & 
\colorbox{orange!100.0}{\strut [CLS]} \colorbox{orange!54.13}{\strut The} \colorbox{orange!26.60}{\strut food} \colorbox{orange!69.21}{\strut is} \colorbox{orange!45.94}{\strut ta} \colorbox{orange!31.88}{\strut \#\#sty} \colorbox{orange!93.54}{\strut but} \colorbox{orange!55.17}{\strut the} \colorbox{orange!32.82}{\strut $<$asp$>$} \colorbox{orange!31.52}{\strut \underline{service}} \colorbox{orange!25.24}{\strut $<$/asp$>$} \colorbox{orange! 73.52}{\strut is} \colorbox{orange! 93.71}{\strut \underline{bad}} \colorbox{orange!90.72}{\strut !} \colorbox{orange! 52.31}{\strut [SEP]} 
\\
\hline
\end{tabular}}
\caption{\textbf{Attention visualization}. Gradient saliency maps~\citep{simonyan2014deep} for the embedding of each word in the transformations under BERT. \underline{Underlined} words are aspects and corresponding opinion spans.}
 \label{visual_attention}    
\end{figure}

\subsection{Visualization of Attention}
To understand the effect of the three transformations, we visualize the attention scores separately offered by our OE model (BERT-MeanPool-Concat) with the transformations, as shown in Figure~\ref{visual_attention}. The four attention vectors have encoded quite different concerns in the token sequence. We can observe that before applying the transformations, the model may attend to more irrelevant words. On the contrary, AC, AP, and AM can promote our model to attend to aspect-specific context and capture the correct opinion spans, thus achieving aspect-specific context modeling in PLM.


\section{Conclusions}
In this paper, we propose three aspect-specific input transformations and methods to leverage these transformations to promote the PLM to pay more attention to the aspect-specific context in two aspect-based sentiment analysis (ABSA) tasks (SC and OE). We conduct experiments  with standard benchmarks for SC and OE, along with adversarial ones for robustness tests. Our models with aspect-specific context modeling achieve the state-of-the-art performance for OE and outperform various strong models for SC. The extensive experimental results and further analysis indicated that aspect-specific context modeling can enhance the performance of ABSA.

\section*{Acknowledgements}
This research was supported in part by Natural Science Foundation of Beijing (grant number:  4222036) and Huawei Technologies (grant number: TC20201228005).

\bibliography{acl_latex}
\bibliographystyle{acl_natbib}

\clearpage
\appendix

\begin{table*}[ht]
\centering
\resizebox{1.0\textwidth}{!}{
\begin{tabular}{lccl}
\hline
\multicolumn{1}{c}{Generation Strategy}                                              & Target Aspect: Opinion     & Other Aspepct:Opinion                               & \multicolumn{1}{c}{Example}                       \\ \hline
\begin{tabular}[c]{@{}l@{}}Source: The original sample from\\ the test set\end{tabular}              
& \begin{tabular}[c]{@{}c@{}}works : well\\ positive\end{tabular} & apple OS : happy                                                                             & Works well , and I am extremely happy to be back to an apple OS .                                                                                                                                            \\
\begin{tabular}[c]{@{}l@{}}RevTgt: Reverse the sentiment of \\ the target aspect\end{tabular}                                                      & \begin{tabular}[c]{@{}c@{}}works : badly\\ negtive\end{tabular} & apple OS : happy                                                                             & Works badly , but I am extremely happy to be back to an apple OS .                                                                                                                                           \\
\begin{tabular}[c]{@{}l@{}}RevNon: Reverse the sentiment of the\\ non-target aspects with  originally \\ the same sentiment as target\end{tabular} & \begin{tabular}[c]{@{}c@{}}works : well\\ positive\end{tabular} & apple OS : unhappy                                                                           & Works well , but I am extremely happy to be back to an apple OS .                                                                                                                                            \\
\begin{tabular}[c]{@{}l@{}}AddDiff: Add aspects with the opposite\\ sentiment from the target aspect\end{tabular}                                  & \begin{tabular}[c]{@{}c@{}}works : well\\ positive\end{tabular} & \begin{tabular}[c]{@{}c@{}}apple OS : happy\\ games : issue\\ video chat : iffy\end{tabular} & \begin{tabular}[c]{@{}l@{}}Works well , and I am extremely happy to be back to an apple OS ,\\ but games being the main issue . And the video chat is the only \\ thing that is iffy about it .\end{tabular} \\ \hline
\end{tabular}}
\caption{The example of using three adversarial strategies to generate the Aspect Robustness Test Set with spans (\textbf{\textsc{ARTS-OE}}) based on \textsc{SemEval}. Specifically, we use these strategies to generate 1002 test instances for the laptop domain (\textbf{\textsc{ARTS-OE-Lap}}) and 2009 test instances for the restaurant domain (\textbf{\textsc{ARTS-OE-Res}}). Each aspect in a sentence is associated with an opinion span for OE.}
\label{ARTS_Span_Example}
\end{table*}

\section{Implementation Details}  \label{section:Implementation_details}
For fair comparison, we re-produce all baselines based on their open-source codes under the same settings. For all the non-PLM models,  300-dimensional GloVe vectors~\citep{pennington2014glove} are leveraged to initialize the input embeddings. All parameters of models are initialized with uniform distributions. The learning rate is 10\textsuperscript{-3}. The coefficient of the L2 regularization is 10\textsuperscript{-5}. In case a model has hidden states, the dimensionality of hidden states is set to 300. For experiments with BERT \citep{devlin2019bert} and RoBERTa \citep{liu2019roberta} as the input embeddings, we adopt the BERT-base-uncased \footnote{\url{https://huggingface.co/bert-base-uncased}.} model and the RoBERTa-base \footnote{\url{https://huggingface.co/roberta-base}.} model as our backbone network respectively, where the dimensionality of hidden states is 768 and the learning rate is set to 10\textsuperscript{-5} for SC and 5*10\textsuperscript{-5} for OE, while the regularization is removed. During all experiments, AdamW \citep{loshchilov2018decoupled} is adopted optimizer in our models. The batch size is 64, and the maximal sequence length is 128. If a model involves attention mechanism, then the dot product-based attention is employed. 

We also carry out experiments on two larger pre-trained models, i.e., BERT-Large and RoBERTa-Large. The experimental results show that the performances are similar to that of BERT-base and RoBERTa-base. Due to space limitation, we do not release the results on BERT-Large and RoBERTa-Large.

It is worth noting that most previous methods did not use the dev set and may have overfitted the test set. We have made a systematic and comprehensive comparison for the first time under the same settings.

\section{Evaluation Metrics} \label{section:Evaluation_metrics}
For standard performance evaluation, each model is trained, validated and tested on the standard datasets for SC and OE. For SC, we use accuracy and macro-averaged  F1-score as performance metrics. Following the previous work \citep{fan2019target}, we adopt F1-score only as the evaluation metric for OE. An opinion extraction is considered correct only when the opinion span predicted is the same as the ground truth. 

To evaluate a model's robustness on SC, the model is trained on the standard \textsc{SemEval} datasets and tested on the corresponding \textsc{ARTS-SC} testsets. For a model' robustness on OE, the model is trained on the standard \textsc{SemEval} datasets and tested on the corresponding \textsc{ARTS-OE} testsets. 

Finally, the experimental results are obtained by averaging five runs with random initialization. It is worth noting that our goal is to verify the effectiveness of the proposed method rather than achieving the sota on SC and OE. Such a simple method can achieve an effectiveness close to sota.

\end{document}